\documentclass{article}
\usepackage{ijcai26}

\usepackage{times}
\usepackage{enumerate}
\usepackage{textcomp}
\usepackage{algorithm}
\usepackage{algorithmic} 
\newcommand{\ignoreText}[1]{}
\usepackage{multirow}
\usepackage{multicol}   
\usepackage{array}
\usepackage{longtable}
\usepackage{verbatim}
\usepackage{tabularx}
\usepackage{amsfonts}
\usepackage{amsmath}
\usepackage{graphicx} 
\usepackage{amssymb}
\usepackage{fancyhdr}
\usepackage{url}
\usepackage{colortbl,hhline} 
\usepackage[margin=15mm]{geometry}
\usepackage{tikz}
\usetikzlibrary{shapes, arrows.meta, positioning}
\usetikzlibrary{patterns}
\usepackage{natbib}
\usepackage{aliascnt}
\usepackage{float}
\usepackage{subcaption}
\title{A Qualitative Model for Reasoning about Path and Support}

\author{
 Abhishek Jaiswal
\and
Zoe Falomir 
\\
\affiliations
Ume{\aa} University, Sweden\\
\emails
\{abhishek.jaiswal,zoe.falomir\}@umu.se
}

\begin{document}

\maketitle

\begin{abstract}

Spatial reasoning abilities correlate strongly with performance in STEM fields.  Games offer a compelling medium for training these critical skills in developing children who have a natural proclivity for play. However, to facilitate human-like tutoring and player guidance, these games require an AI agent capable of making commonsense inferences from spatial events. Qualitative reasoning (QR) models appear to be a suitable framework for these application domains. As these models reason in symbolic representations, they can seamlessly translate game states into interpretable feedback for human-like player guidance. This paper introduces a hybrid qualitative model designed for Camelot Jr., a block-puzzle game that requires constructing multi-level bridges to connect two avatars stationed on separate towers.  The game poses a challenge for the player, who must make platforms stable, plan their path, and ensure they use all the provided blocks. To handle the precise physics required by the domain, we integrate a mathematical center-of-mass stability logic to guide our qualitative solver. Our work facilitates spatial skill training in Camelot Jr. and contributes to the development of human-centric, explainable game-playing agents.

\end{abstract}

\section{Introduction}
\label{sec:intro}

Qualitative Spatial and Temporal Representations and Reasoning (QSTR)  \citep{Cohn_Renz:2007, Ligozat_2011, sioutis2021qualitative} models and reasons about
\textit{time} (i.e., coincidence, order, concurrency, overlap, granularity) and also about 
properties of \textit{space} (i.e., topology, location, direction, proximity, geometry, intersection, etc.) 
as well as their evolution in time across continuous neighboring situations.
Maintaining spatial and temporal consistency and constraints is fundamental to qualitative reasoning
when solving spatial problems (such as pathfinding, orientation, and relative positioning) and temporal problems (such as constraint satisfaction, schedule optimization, and precedence)~ \citep{Proc-IJCAI:2009}.
As a result, well-defined qualitative models and reasoning techniques have been invented in the literature that can handle imprecise and incomplete knowledge at the symbolic level.
Spatio-temporal reasoning has proven successful across a wide range of domains and applications such as:  robotics \citep{jetai:MastFW16,Zfalomir_JSCC:2011,Falomir_PRL:2013}, computer vision
\citep{Suchan:PhD}, ambient intelligence \citep{NeurocomFalomir:2015}, architecture and design \citep{LiSchultz24,MehulFreksa:2015}, geographic information systems
\citep{gis/WeiGC24,ClementiniC24,abs-2512-15388}, 
safety analysis \citep{GrigoleitHPRRSW17}, autonomous driving \citep{Bhat_Suchan_Mosen_ASP_AutDriving:2025}, game playing \citep{jaiswal2024learning} and education \citep{KragtenHB24,SantamartaGVMF23}. 
A comprehensive survey of the most relevant calculi in Qualitative Spatial Reasoning (QSR) can be found in \citet{DyllaL-Survey:2014}.


Furthermore, qualitative representations are believed to be closer to the cognitive domain, as shown in cognitive models of sketch recognition \citep{Lovett}, spatial problem-solving tasks (e.g., visual oddity tasks) \citep{Lovett-Cognition:2011}, paper-folding reasoning tests \citep{Falomir-QR:2016,mta/FalomirTPG21}, analogy in Raven’s Progressive Matrices \citep{Lovett-Forbus:2017}, logic composition of qualitative shapes in solving cognitive tests \citep{FalomirPC20,PichF18}, 
modeling 2D spatial abstraction during mental rotation tasks \citep{Lovett-CogSci:2014}, solving 3D rotation tasks \citep{Ejarque_Falomir:QR:2026}, and
solving the Cube Comparison Test (CCT)~\citep{SpatialCognition:2026,Falomir:QOR:arxiv}.

Therefore, novel models that combine QSTR, cognitive spatial thinking, and common sense are a challenge that holds the promise of further advances in Artificial Intelligence and its applications.

Moreover, spatial cognition studies 
have shown that there is a strong link between spatial abilities and success in Science, Technology, Engineering and Math (STEM) disciplines \citep{Newcombe:2010,Wai_et_al:2009}.
For example, children as young as four years  old  already have an informal awareness of spatial relations, such as identifying parallel lines for two dimensional shape description, well before they are formally taught about them \citep{Sinclair-et-al:2013}.
For this reason, researchers in the US and Canada study the current status and future possibilities of spatial reasoning in contemporary school mathematics \citep{Sinclair-Bruce:2014}, also because spatial learning and reasoning can be easily taught using the visual and kinetic interactions offered by new digital technologies \citep{Highfield:2007}. 


In cognitive psychology, games like \textit{Upside Down World} are used to evaluate spatial skills of students by challenging them to recreate buildings composed of multilink cubes in their upright orientation and use spatial language to describe the composition of the buildings to their colleagues for building them accordingly \citep{Sinclair-Bruce:2014}.
Also, to help identify gifted children, the German Academic Foundation tests kids on matching the consistent projection of a 3D object corresponding to a technical drawing\footnote{\textit{Test der Studienstiftung: Gehirnjogging f{\"{u}}r Hochbegabte}, see \textit{Spiegel} Online: \url{http://www.spiegel.de/quiztool/quiztool-49771.html}}. In previous works by one of the authors, a qualitative model for 3D object description was developed, and promising results were obtained \citep{FedCSIS2015370}.

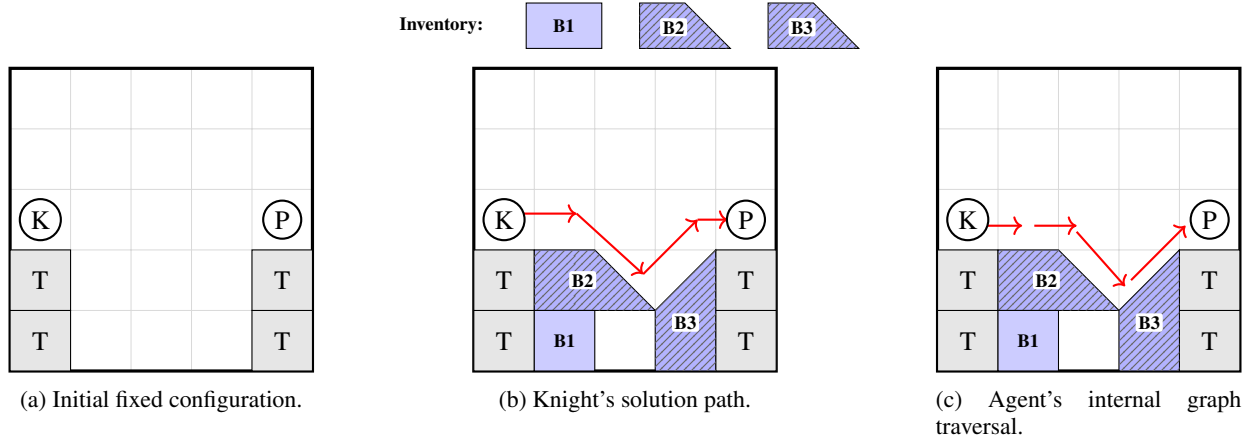
\begin{figure*}[t!] 
\centering

    \begin{minipage}{\textwidth}
        \centering
        \begin{tikzpicture}[scale=1]
            \node[anchor=west, font=\scriptsize] at (-3., 0.3) {\textbf{Inventory:}};
            \draw[fill=blue!20] (-1.2, 0) rectangle (-0.2, 0.6) node[pos=.5] {\scriptsize \textbf{B1}};
            \draw[fill=blue!30, postaction={pattern=north east lines, pattern color=black!60}] 
                (0.3, 0) -- (1.5, 0) -- (0.9, 0.6) -- (0.3, 0.6) -- cycle;
            \node[fill=white, inner sep=1pt, rounded corners=1pt, font=\scriptsize] at (0.75, 0.28) {\textbf{B2}};
            \draw[fill=blue!30, postaction={pattern=north east lines, pattern color=black!60}] 
                (2.0, 0) -- (3.2, 0) -- (2.6, 0.6) -- (2.0, 0.6) -- cycle;
            \node[fill=white, inner sep=1pt, rounded corners=1pt, font=\scriptsize] at (2.45, 0.28) {\textbf{B3}};
        \end{tikzpicture}
    \end{minipage}

    \vspace{0.2cm} 

    \begin{minipage}[b]{0.32\textwidth}
        \centering
        \subcaptionbox{Initial fixed configuration.\label{fig:camelot_initial}}{%
            \begin{tikzpicture}[scale=0.8]
                \draw [gray!30, very thin] (0, 0) grid (5, 5);
                \draw[very thick] (0, 0) rectangle (5, 5);
                
                \node[fill=gray!20, minimum size=0.8cm, draw] at (0.5, 0.5) {T};
                \node[fill=gray!20, minimum size=0.8cm, draw] at (0.5, 1.5) {T};
                \node[fill=gray!20, minimum size=0.8cm, draw] at (4.5, 0.5) {T};
                \node[fill=gray!20, minimum size=0.8cm, draw] at (4.5, 1.5) {T};
            
                \node[draw, circle, fill=white, inner sep=2pt, thick] at (0.5, 2.5) {K};
                \node[draw, circle, fill=white, inner sep=2pt, thick] at (4.5, 2.5) {P};
            \end{tikzpicture}%
        }
    \end{minipage}%
    \hspace{0.01\textwidth}%
    \begin{minipage}[b]{0.32\textwidth}
        \centering
        \subcaptionbox{Knight's solution path.\label{fig:camelot_placed2}}{%
            \begin{tikzpicture}[scale=0.8]
                \draw [gray!30, very thin] (0, 0) grid (5, 5);
                \draw[very thick] (0, 0) rectangle (5, 5);
                
                \node[fill=gray!20, minimum size=0.8cm, draw] at (0.5, 0.5) {T};
                \node[fill=gray!20, minimum size=0.8cm, draw] at (0.5, 1.5) {T};
                \node[fill=gray!20, minimum size=0.8cm, draw] at (4.5, 0.5) {T};
                \node[fill=gray!20, minimum size=0.8cm, draw] at (4.5, 1.5) {T};
                
                \draw[fill=blue!30, postaction={pattern=north east lines, pattern color=black!60}] 
                    (1, 1) -- (3, 1) -- (2, 2) -- (1, 2) -- cycle;
                \node[fill=white, inner sep=1pt, rounded corners=1pt, font=\scriptsize] at (1.8, 1.5) {\textbf{B2}};
                
                \draw[fill=blue!30, postaction={pattern=north east lines, pattern color=black!60}] 
                    (3, 0) -- (4, 0) -- (4, 2) -- (3, 1) -- cycle;
                \node[fill=white, inner sep=1pt, rounded corners=1pt, font=\scriptsize] at (3.5, .8) {\textbf{B3}};
                \draw[fill=blue!20] (1, 0) rectangle (2, 1) node[pos=.5] {\scriptsize \textbf{B1}};

                \draw[->, red, thick] (0.7, 2.6) -- (1.7, 2.6);
                \draw[->, red, thick] (1.7, 2.6) -- (2.8, 1.6);
                \draw[->, red, thick] (2.8, 1.6) -- (3.7, 2.5);
                \draw[->, red, thick] (3.7, 2.5) -- (4.2, 2.5);
                
                \node[draw, circle, fill=white, inner sep=2pt, thick] at (0.5, 2.5) {K};
                \node[draw, circle, fill=white, inner sep=2pt, thick] at (4.5, 2.5) {P};
            \end{tikzpicture}%
        }
    \end{minipage}%
    \hspace{0.01\textwidth}%
    \begin{minipage}[b]{0.32\textwidth}
        \centering
        \subcaptionbox{Agent's internal graph traversal.\label{fig:camelot_path3}}{%
            \begin{tikzpicture}[scale=0.8]
                \draw [gray!30, very thin] (0, 0) grid (5, 5);
                \draw[very thick] (0, 0) rectangle (5, 5);
                
                \node[fill=gray!20, minimum size=0.8cm, draw] at (0.5, 0.5) {T};
                \node[fill=gray!20, minimum size=0.8cm, draw] at (0.5, 1.5) {T};
                \node[fill=gray!20, minimum size=0.8cm, draw] at (4.5, 0.5) {T};
                \node[fill=gray!20, minimum size=0.8cm, draw] at (4.5, 1.5) {T};
                
                \draw[fill=blue!30, postaction={pattern=north east lines, pattern color=black!60}] 
                    (1, 1) -- (3, 1) -- (2, 2) -- (1, 2) -- cycle;
                \node[fill=white, inner sep=1pt, rounded corners=1pt, font=\scriptsize] at (1.8, 1.5) {\textbf{B2}};
                
                \draw[fill=blue!30, postaction={pattern=north east lines, pattern color=black!60}] 
                    (3, 0) -- (4, 0) -- (4, 2) -- (3, 1) -- cycle;
                \node[fill=white, inner sep=1pt, rounded corners=1pt, font=\scriptsize] at (3.5, .8) {\textbf{B3}};
                \draw[fill=blue!20] (1, 0) rectangle (2, 1) node[pos=.5] {\scriptsize \textbf{B1}};

                \draw[->, red, thick] (0.7, 2.4) -- (1.4, 2.4);
                \draw[->, red, thick] (1.6, 2.4) -- (2.3, 2.4);
                \draw[->, red, thick] (2.3, 2.3) -- (3.1, 1.4);
                \draw[->, red, thick] (3.2, 1.5) -- (4.1, 2.4);
                
                \node[draw, circle, fill=white, inner sep=2pt, thick] at (0.5, 2.5) {K};
                \node[draw, circle, fill=white, inner sep=2pt, thick] at (4.5, 2.5) {P};
            \end{tikzpicture}%
        }
    \end{minipage}

\caption{Qualitative discrete representation of a \textit{Camelot Jr.} game challenge layout: (a) initial problem state, (b) a stable solution path with stair-blocks placed in horizontal and vertical orientations, and (c) the agent's internal path traversal graph. The agent's moves are: (\textit{move-right}), (\textit{move-right} + \textit{move-diagonal-down}), and (\textit{move-diagonal-up}).}
\label{fig:camelot_combined_master}
\end{figure*}

Consequently, QSR methods present a unique opportunity to develop game-playing agents that follow a human-centric logic. While this paradigm requires frameworks grounded in symbolic representations, existing research on spatial games like \textit{Angry Birds} suggests the need to complement pure symbolic abstractions by mathematical knowledge of physical phenomena~\cite{haase2021behind}. In this work, we investigate a hybrid spatial reasoning framework using the bridge-building puzzle game \textit{Camelot Jr.} (illustrated in Figure~\ref{fig:camelot_combined_master}), where the core objective is to construct a continuous path connecting a starting position to a goal point. To address this, we propose a solver model grounded in human-like commonsense logic that combines qualitative reasoning modules with physical stability calculations.

Concretely, this paper explores the challenge of formulating a model that can serve two objectives: 
\begin{itemize} 

\item Help people understand how to solve games involving spatial stability  and path planning, so that they can improve their spatial skills and therefore enhance their success in STEM; and also,

\item Equip artificial intelligence agents with the symbolic and physical reasoning mechanisms to solve spatial problems in wooden puzzle games so that they can reason about the spatial stability and path finding.

\end{itemize}

The rest of the paper is organized as follows. 
Section \ref{sec:game} outlines the specifics of the Camelot Jr. game. 
Section \ref{sec:model} introduces the Qualitative Descriptor model for the game.
Section \ref{sec:BlockPlace} describes the game constraints for the qualitative model to comply with.
Section \ref{sec:support} details the numerical centre-of-mass (COM) stability logic.
Section \ref{sec:cell_walk} describes the graph traversal logic used by the agent in the game.
Section \ref{sec:path_finding} explains the path-finding algorithm using the previously discussed ingredients.
Section \ref{sec:discussion} concludes with a discussion on the current scope and future directions.


\section{The Camelot Jr. Game: Background}

This section presents the various game elements, how they interact, and the rules governing their behavior.
\label{sec:game}
\subsection{Game Elements and Definitions}
All gameplay in \textit{Camelot Jr.} occurs on a vertical 2D \textbf{Grid} that represents the castle layout, where the \textbf{Ground} serves as the base of the grid and acts as the primary immutable support for all structures. The core task revolves around two animated characters, the \textbf{Knight} ($K$) and the \textbf{Princess} ($P$), who must be connected by a path comprising various block pieces across the layout. The environment contains fixed \textbf{Towers} ($T$), which are solid, permanent structures positioned at the start of a challenge. To bridge structural gaps, the player deploys pieces from a limited inventory of \textbf{Blocks} ($B$), consisting of rectangular building blocks that serve as stable, walkable platforms and stair-blocks, which combine a solid block with a ramp side that allows diagonal movement up or down. The ramp side occupies exactly one grid cell. The rectangular blocks can be of sizes 1 to 3. The stair-block has one ramp side combined with rectangular blocks of size one or two (see Figure~\ref{fig:orientations} for sample stair-block orientations).

The entire environment is governed by a global \textbf{Gravity} constraint, ensuring that every placed block structure is physically viable under static equilibrium to be considered a valid solution. Together, these elements dictate the available \textbf{Walkable Space}, defined as the set of empty grid cells located immediately above stable structures or the ground. Crucially, for stair-blocks, the ramp face itself constitutes the active walkable surface rather than the empty grid cell above it. A schematic overview of the game is illustrated in Figure~\ref{fig:camelot_combined_master}.

 \subsection{Game Mechanics and Objectives}
The primary objective in \textit{Camelot Jr.} is to use a limited set of blocks to construct a path for the Knight to reach the Princess. This task is governed by three environmental constraints. First, no game element can extend beyond the castle grid. Second, towers stay fixed; they cannot be moved or rotated. Third, every inventory block given at the start must be used in the final solution.

Beyond these operational boundaries, the player must follow the rules of structural placement and stability. The support rule requires that every block must rest on solid ground, a fixed tower, or another stable block. Under the global stability rule, any assembly of inventory blocks must remain viable under gravitational torque without toppling. The inclined side of a stair-block cannot support another block (i.e., no block can be placed on top of the ramp for stability). However, if a horizontal stair-block is inverted with its flat bottom on top (\textit{HNI} and \textit{HMI} in Figure~\ref{fig:orientations}), it provides a valid horizontal support surface. A valid solution gives a walkable path, defined as a sequence of connected grid cells, from the Knight to the Princess.

\begin{table*}[t] 
\centering
\begin{tabular}{|l|l|c|l|}
\hline
\textbf{Label} & \textbf{State Name} & $(\text{Axis}, f, \iota) \text{ if } \chi=1$ & \textbf{Functional Affordance} \\ \hline
\textbf{HN} & Horizontal Normal & $(H, 0, 0)$ & Flat structural support + Downward ramp from left to right \\ \hline
\textbf{HM} & Horizontal Mirror & $(H, 1, 0)$ & Flat structural support + Upward ramp from left to right \\ \hline
\textbf{HNI} & Horizontal Inverted & $(H, 0, 1)$ & Flat structural support \\ \hline
\textbf{HMI} & Horizontal Mirror Inverted & $(H, 1, 1)$ & Flat structural support \\ \hline
\textbf{VN} & Vertical Normal & $(V, 0, 0)$ & Upward ramp from left to right \\ \hline
\textbf{VM} & Vertical Mirror & $(V, 1, 0)$ & Downward ramp from left to right \\ \hline
\textbf{VNI} & Vertical Normal Inverted & $(V, 0, 1)$ & Inverted Pillar (Illegal placement) \\ \hline
\textbf{VMI} & Vertical Mirror Inverted & $(V, 1, 1)$ & Inverted Pillar (Illegal placement) \\ \hline \hline
\textbf{RB} & Rectangular Solid Block & $\chi=0$ & Flat structural support \\ \hline
\end{tabular}
\caption{Qualitative orientation labels and the supported affordances for inventory blocks}
\label{tab:orientation_lexicon}
\end{table*}

\section{Qualitative Description}
\label{sec:model}

\subsection{ Grid Environment ($\mathcal{G}$)}
The grid is a set of cells $c_{x,y}$ where $x \in [0, W)$ and $y \in [0, H)$:
\[ c_{x,y} = \begin{cases} 0 & \text{Empty Space} \\ 1 & \text{Tower } (T) \\ id \ge 1 & \text{Placed Block } (B) \end{cases} \]

 $K = (x_k, y_k)$ (Knight) and $P = (x_p, y_p)$ (Princess).

\subsection{Block Definition}
A block $B$ is defined as a 4-tuple representing its intrinsic properties and current state:

\[ B = \{N, \text{id}, \mathcal{O}, \chi\} \]

Where:
\begin{itemize}
    \item $N \in \mathbb{Z}^+$ is the \textbf{Scalar Size}, representing the length of the block segment.
    \item \text{id} $\in \mathbb{Z}$ is the unique identifier used for tracking grid occupancy and enforcing placement order.
    \item $\mathcal{O}$ is the \textbf{Orientation State Vector}, stores the particular orientation of the block to be used for grid placement.
    \item $\chi \in \{0, 1\}$ is the \textbf{Stair Indicator}, defining whether the block possesses ramp side ($\chi=1$) or is a standard flat block ($\chi=0$). 
\end{itemize}
A placement is a mapping $L: B \to (x, y)$, where $(x, y)$ represents the anchor coordinate (the bottom-left cell) of the block in the grid $\mathcal{G}$. We define $B_{x,y}$ as the block instance whose anchor point is situated at $(x,y)$

\subsection{The Orientation State Vector}

The vector $\mathcal{O} = (\text{Axis}, f, \iota)$ governs the spatial projection and parity of the block via three qualitative operations:

\begin{itemize}
\item \textbf{Axis} $\in \{H, V\}$: Determines if the block is Horizontal or Vertical.
\item \textbf{Reflection} ($f \in \{0, 1\}$): Governs horizontal parity (Mirroring).
\item \textbf{Inversion} ($\iota \in \{0, 1\}$): Governs vertical parity (Flipping).
\end{itemize}

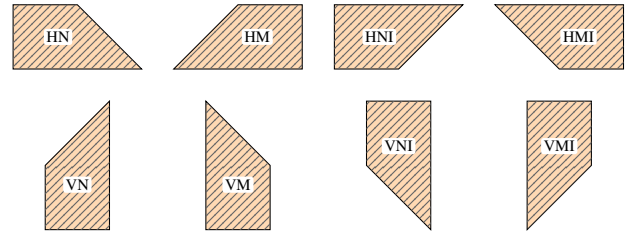
\begin{figure}[t] 
\centering
\begin{tikzpicture}[scale=0.85]


\begin{scope}[shift={(0,0)}]
    \draw[fill=orange!30, postaction={pattern=north east lines, pattern color=black!60}] 
        (0,0) -- (2,0) -- (1,1) -- (0,1) -- cycle;
    \node[fill=white, inner sep=1pt, rounded corners=1pt] at (0.7, 0.5) {\tiny HN};
\end{scope}

\begin{scope}[shift={(2.5,0)}]
    \draw[fill=orange!30, postaction={pattern=north east lines, pattern color=black!60}] 
        (0,0) -- (2,0) -- (2,1) -- (1,1) -- cycle;
    \node[fill=white, inner sep=1pt, rounded corners=1pt] at (1.3, 0.5) {\tiny HM};
\end{scope}

\begin{scope}[shift={(5,0)}]
    \draw[fill=orange!30, postaction={pattern=north east lines, pattern color=black!60}] 
        (0,0) -- (1,0) -- (2,1) -- (0,1) -- cycle;
    \node[fill=white, inner sep=1pt, rounded corners=1pt] at (0.7, 0.5) {\tiny HNI};
\end{scope}

\begin{scope}[shift={(7.5,0)}]
    \draw[fill=orange!30, postaction={pattern=north east lines, pattern color=black!60}] 
        (1,0) -- (2,0) -- (2,1) -- (0,1) -- cycle;
    \node[fill=white, inner sep=1pt, rounded corners=1pt] at (1.3, 0.5) {\tiny HMI};
\end{scope}


\begin{scope}[shift={(0.5,-2.5)}]
    \draw[fill=orange!30, postaction={pattern=north east lines, pattern color=black!60}] 
        (0,0) -- (1,0) -- (1,2) -- (0,1) -- cycle;
    \node[fill=white, inner sep=1pt, rounded corners=1pt] at (0.5, 0.7) {\tiny VN};
\end{scope}

\begin{scope}[shift={(3,-2.5)}]
    \draw[fill=orange!30, postaction={pattern=north east lines, pattern color=black!60}] 
        (0,0) -- (1,0) -- (1,1) -- (0,2) -- cycle;
    \node[fill=white, inner sep=1pt, rounded corners=1pt] at (0.5, 0.7) {\tiny VM};
\end{scope}

\begin{scope}[shift={(5.5,-2.5)}]
    \draw[fill=orange!30, postaction={pattern=north east lines, pattern color=black!60}] 
        (0,1) -- (1,0) -- (1,2) -- (0,2) -- cycle;
    \node[fill=white, inner sep=1pt, rounded corners=1pt] at (0.5, 1.3) {\tiny VNI};
\end{scope}

\begin{scope}[shift={(8.0,-2.5)}]
    \draw[fill=orange!30, postaction={pattern=north east lines, pattern color=black!60}] 
        (0,0) -- (1,1) -- (1,2) -- (0,2) -- cycle;
    \node[fill=white, inner sep=1pt, rounded corners=1pt] at (0.5, 1.3) {\tiny VMI};
\end{scope}

\end{tikzpicture}
\caption{The eight qualitative orientation states for a stair-block ($N>=2$). Arrangements like VNI and VMI are illegal as per the game rules.}
\label{fig:orientations}
\end{figure}

The solver navigates the orientation space using three discrete operations: 
\begin{itemize}
\item \textbf{Reflect ($R$):} A Horizontal Flip (Reflection across the vertical axis), 
\item \textbf{Invert ($I$):} A Vertical Flip (Reflection across the horizontal axis), 
\item \textbf{Transpose ($T$):} A $90^\circ$ Anti-Clockwise Rotation.
\end{itemize}
Table~\ref{tab:orientation_lexicon} describes the various orientations for a stair-block, along with the affordance of the orientations. Figure~\ref{fig:orientations} shows a schematic of the various ramp orientations. Horizontal Normal refers to the base orientation of a stair-block with the ramp occupying the rightmost size (1st diagram in Figure~\ref{fig:orientations}).

For a rectangular block denoted as \textbf{RB}, only two distinct orientations exist - Horizontal and Vertical and we use only the Axis label to define them.

\subsection{State Transition Matrix} Table~\ref{tab:transition_matrix} defines how orientation changes through various qualitative operations for a stair-block. The solver utilizes this to explore valid placements during backtracking. For rectangular blocks only to-and-fro Horizontal-Vertical state transitions exist. From subsequent discussion, we consider only the legal placements of stair-blocks.

\begin{table}[h]
\centering
\renewcommand{\arraystretch}{1.2} 
\begin{tabular}{|l||c|c|c|}
\hline
\small \textbf{Initial State} & \small \textbf{Reflect ($R$)} & \small \textbf{Invert ($I$)} & \small \textbf{Transpose ($T$)} \\ \hline\hline
HN & HM & HNI & VN \\ \hline
HM & HN & HMI & VNI (Illegal) \\ \hline
HNI & HMI & HN & VM \\ \hline
HMI & HNI & HM & VMI (Illegal)\\ \hline
VN & VM & VNI (Illegal) & HMI \\ \hline
VM & VN & VMI (Illegal) & HM \\ \hline
\end{tabular}
\caption{State Transition Matrix for qualitative block orientations. Pruned states are marked as Illegal.}
\label{tab:transition_matrix}
\end{table}


\section{Block Placement Constraints for the Solver}
\label{sec:BlockPlace}
\subsection{Spatial Restrictions}
The effective dimensions $(w_e, h_e)$ are emergent properties derived from the Scalar Size $N$ and the \textbf{Axis} state:
\[ (w_e, h_e) =
\begin{cases}
(N, 1) & \text{if Axis} = H \\
(1, N) & \text{if Axis} = V
\end{cases} 
\]
A placement $L: B \to (x, y)$ is valid only if every grid cell $c$ within the block's effective dimensions $(w_e, h_e)$ is currently empty ($\text{value} = 0$).
$$ \forall j \in [0, w_e), \forall i \in [0, h_e): c_{x+j, y+i} = 0 $$
This rule ensures that blocks do not collide with fixed Towers or previously placed inventory items.
Additionally, the placement must satisfy the Boundary Constraint:$$ x + w < W \quad \text{and} \quad y + h < H $$

\subsection{Foundational Restrictions}



To optimize the search process, the solver evaluates a block's immediate foundational support before committing to a placement. For any block $B_{x, y}$ not at ground level ($y > 0$), there must be at least one valid point of support contact with the structure below. 
Support is validated by inspecting the cells directly beneath the block's footprint. For these supporting blocks, we have:

 \begin{flalign*}
    &\text{For } j \in [0, w_e] \text{ and } (x_{\text{sup}}, y_{\text{sup}}) \in c_{x+j, y-1}, &\\
    &\quad \quad \quad \quad \quad \text{ValidSupport}(x_{\text{sup}}, y_{\text{sup}}) = \text{True if:} &
\end{flalign*}

\begin{enumerate}
    \item $c_{x_{\text{sup}},y_{\text{sup}}} \neq 0$ i.e., $x_{\text{sup}},y_{\text{sup}}$ must be part of a Block $B_{\text{sup}}$
    \item $\chi=0$ i.e., $B_{\text{sup}}$ is not a stair-block.
    \item If $B_{\text{sup}} \in \{\text{HNI}, \text{HMI}\}$ (i.e., $\chi=1$ and $\iota=1$), it is an inverted stair-block providing flat structural support.
    \item If $B_{\text{sup}} \in \{\text{HN}, \text{HM}, \text{VN}, \text{VM}\}$ (i.e., $\chi=1$ and $\iota=0$), the support is invalid if $(x_{\text{sup}},y_{\text{sup}})$ is the non-inverted ramp cell, i.e.,\\
    {\small
    $(x_{\text{sup}},y_{\text{sup}}) \neq \begin{cases} 
        (x_s + N - 1, y_s) & \text{if } \text{label}(B_{\text{sup}}) = \text{HN} \\
        (x_s, y_s)         & \text{if } \text{label}(B_{\text{sup}}) = \text{HM} \\
        (x_s, y_s + N - 1) & \text{if } \text{label}(B_{\text{sup}}) \in \{\text{VN}, \text{VM}\} 
    \end{cases}$%
    }
\end{enumerate}
\noindent where $(x_s, y_s)$ denotes the structural anchor (the bottom-left reference cell) of the supporting block $B_{\text{sup}}$, and $N$ represents the dimension length of the block.

This support logic works because any non-stair horizontal or vertical block allows walking over itself. Similarly, if the stair is inverted ($\iota=1$), the block acts as a flat top surface, and support is valid throughout its width. For a normal stair-block, the support is valid as long as the cell $x_{sup},y_{sup}$ is not the ramp cell of the stair-block.

\section{Recursive Stability Formalization}
\label{sec:support}

The solver verifies structural integrity by iterating through all placed pieces in descending order of their vertical position ($y$). This top-down sequence follows real-world centre-of-mass (COM) physics and ensures that loads from higher blocks are propagated to those below.

\subsection*{Recursive Load Aggregation}
From top to bottom in each stack of blocks, for each block $B_i$ with weight $w_i$, we define the total weight $W_i$ from the top till $B_i$ including it and their resultant moment $M_i$:
\[ W_i = w_i + \sum_{k \in \text{Loads}(i)} w_k \]
\[ M_i = (w_i \cdot c_i) + \sum_{k \in \text{Loads}(i)} (w_k \cdot c_k) \]
Here  $Loads(i)$ represents all blocks above $B_i$, $c_i$ and $c_k$ represent the COM for $B_i$ and blocks above it.

The effective COM for the system including $B_i$ is then:
\[ COM_{res,i} = \frac{M_i}{W_i} \]

\subsection*{The Convex Hull Constraint}
A sub-stack of blocks is considered stable if, and only if, its resultant center of mass is contained within the bounds of its support geometry:
\[ s_{min} \le COM_{res,i} \le s_{max} \]
Where $s_{min}$ and $s_{max}$ represent the extreme horizontal boundaries of all supporting cells $(x, y-1)$ in contact with the base block ($B_i$) of the current sub-stack.

\subsection*{Bridge Distribution (The Lever Rule)}
In cases of multiple supports (e.g., a bridge), the total weight $W_i$ is distributed between the two outermost supports $s_1$ and $s_2$ with horizontal center points $cp_1$ and $cp_2$ (where $cp_1$  \textless{} $cp_2$). The force $f_1$ allocated to support $s_1$ is determined by:
\[ f_1 = W_i \cdot \frac{cp_2 - COM_{res,i} }{cp_2 - cp_1} \]
 The remaining force $f_2 = W_i - f_1$ is transmitted to $s_2$. This recursive propagation continues until the forces are absorbed by the immutable ground or towers.

\section{Cell Affordances for Agent Actions}
\label{sec:cell_walk}


Once the entire block inventory is placed on the grid and verified as structurally stable, the agent attempts to discover a valid path connecting the Knight to the Princess. 
Starting from the Knight's position, the agent operates within a Directed Acyclic Graph (DAG) framework, with the movement primitives \textit{move-right}, \textit{move-diagonal-up}, and \textit{move-diagonal-down} (Table~\ref{tab:actions}). The agent initiates a  \textit{move-right} motion from an idle position; however, depending on the structural affordances of the terrain where the agent lands, a \textit{move-diagonal-up} or \textit{move-diagonal-down} adjustment may follow automatically depending on where the agent lands.
\begin{table}[h!]
\centering
    \begin{tabular}{ll}
    \hline
    \textbf{Action Label} & \textbf{Coordinate Transition} \\ \hline
    \textit{move-diagonal-up}   & $(i, j) \to (i+1, j+1)$ \\
    \textit{move-diagonal-down} & $(i, j) \to (i+1, j-1)$ \\
    \textit{move-right}         & $(i, j) \to (i+1, j)$   \\ \hline
    \end{tabular}
    \caption{Permissible agent actions during path exploration from the Knight to the Princess.}
\label{tab:actions}
\end{table}

Let the intentional horizontal target cell be $c_{x,y}$ at grid position $(x, y) = (i+1, j)$, and let $(x_{\text{sup}}, y_{\text{sup}}) = (i+1, j-1)$ be the underlying support cell provided by block $B_{\text{sup}}$. Let $(x_s, y_s)$ denote the bottom-left anchor point of the block under evaluation (i.e., the support block if the target cell is empty, or the block occupying the target cell otherwise). A transition is valid if it satisfies any of the following conditions:

\begin{enumerate}
    \item \textbf{Flat Foundation ($\Delta y = 0$):} A horizontal move is valid if the cell is empty and the agent is supported by a flat boundary. This occurs if the support cell directly below is ground ($j-1=0$), a tower, or a block with flat structural support. For inventory block support, we require: 
    $\begin{aligned}[t]
    &c_{x,y} = 0 \quad \land
    &c_{x_{\text{sup}}, y_{\text{sup}}} \in \{\text{RB}, \text{HNI}, \text{HMI}\}
    \end{aligned}$
    
    \item \textbf{Stair Platform ($\Delta y = 0$):} A horizontal move is still valid if the agent walks along the flat surface of an upright stair-block. Only $\text{HN}$ and $\text{HM}$ blocks present this specific combination of a flat walking surface and a ramp cell:
    $\begin{aligned}[t]
    &c_{x,y} = 0 \quad \land \quad c_{x_{\text{sup}}, y_{\text{sup}}} \in \{\text{HN}, \text{HM}\} \text{ such that} \\
    &(x_{\text{sup}}, y_{\text{sup}}) \neq \begin{cases} 
        (x_s + N - 1, y_s) & \text{if } \text{label}(B_{\text{sup}}) = \text{HN} \\
        (x_s, y_s)         & \text{if } \text{label}(B_{\text{sup}}) = \text{HM} 
    \end{cases}
    \end{aligned}$
    
    \item \textbf{Geometric Descent ($\Delta y = -1$):} An automatic diagonal downward transition is triggered when the horizontal movement causes the agent to land over an empty cell whose underlying support footprint corresponds exactly to a downward-sloping ramp cell.
    $\begin{aligned}[t]
    &c_{x,y} = 0 \quad \land \quad c_{x_{\text{sup}}, y_{\text{sup}}} \in \{\text{HN}, \text{VM}\} \text{ such that} \\
    &(x_{\text{sup}}, y_{\text{sup}}) = \begin{cases} 
        (x_s + N - 1, y_s) & \text{if } \text{label}(B_{\text{sup}}) = \text{HN} \\
        (x_s, y_s + N - 1) & \text{if } \text{label}(B_{\text{sup}}) = \text{VM} 
    \end{cases}
    \end{aligned}$
    
    \item \textbf{Geometric Ascent ($\Delta y = +1$):} A diagonal upward transition automatically follows if the agent lands on the ramp cell of an ascending stair. This is the only case in which the landing cell for \textit{move-right} is not an empty cell ($c_{x,y} \neq 0$), and validity requires the entry point to match the ramp cell.
    $\begin{aligned}[t]
    &c_{x,y} \neq 0 \quad \land \quad c_{x,y} \in \{\text{HM}, \text{VN}\} \text{ such that} \\
    &(x, y) = \begin{cases}  
        (x_s, y_s)         & \text{if } \text{label}(B_{x,y}) = \text{HM} \\
        (x_s, y_s + N - 1) & \text{if } \text{label}(B_{x,y}) = \text{VN} 
    \end{cases}
    \end{aligned}$
\end{enumerate}

Figure~\ref{fig:camelot_path3} illustrates the agent's motion as it executes two consecutive \textit{move-right} actions from the Knight's initial position. It lands on a cell above the downward ramp, which automatically triggers a \textit{move-diagonal-down} action. Subsequently, it lands on an upward ramp, triggering a \textit{move-diagonal-up} action and enabling the agent to successfully reach the Princess. Although the agent's path differs slightly from the Knight's traced path due to stair-block traversal mechanics, both approaches follow the same block-placement structure. 
\section{Pathfinding Agent Behaviour}
\label{sec:path_finding}
The agent's solving algorithm is implemented using a recursive backtracking protocol. Algorithm \ref{alg:backtracking_solver} describes the step-by-step working of the agent. It searches the permutation space of the inventory $\mathcal{I}$ to identify a final structural arrangement that satisfies two independent evaluation criteria: \textit{Static Stability} and \textit{Path Connectivity}.

Mirroring a human player, the agent places inventory blocks one after another onto the grid layout, evaluates their structural validity once the inventory is fully deployed, and subsequently evaluates path connectivity. 
Similar to an intelligent player who intuitively rules out any piece configuration unlikely to reach the target destination, our solver implements geometric pruning strategies to cut off impractical search branches that cannot connect the start and the end points with the remaining inventory.

\begin{algorithm}[t]
\caption{Recursive Backtracking Search for Stable Path Solutions}
\label{alg:backtracking_solver}
\begin{algorithmic}[1]
\REQUIRE Inventory $\mathcal{I}$, Current grid configuration $\mathcal{G}$, Placed blocks set $\mathcal{U}$
\ENSURE A stable grid configuration with a valid path, or failure

\IF{$|\mathcal{U}| = |\mathcal{I}|$}
    \STATE \COMMENT{Base Case: Inventory deployed}    
    \IF{\textbf{not} IsEntireStructureStable$(\mathcal{G})$}

        \RETURN \textbf{false} \COMMENT{Reject: unstable structure}
    \ENDIF
    \STATE $\textit{Path} \gets$ FindValidPath$(\mathcal{G})$
    \IF{$\textit{Path} \neq \emptyset$}
        \STATE SaveSolution$(\mathcal{G}, \textit{Path})$
        \RETURN \textbf{true} \COMMENT{Solution confirmed}
    \ENDIF
    \RETURN \textbf{false} \COMMENT{Reject: unreachable goal}
\ENDIF

\STATE $B \gets$ SelectUnusedBlock$(\mathcal{I} \setminus \mathcal{U})$
\FOR{each Orientation $O \in$ GetOrientations$(B)$}
\FOR{each Position $(x, y) \in$ GridCoordinates}
    \IF{\textbf{not} HasValidSupportFootprint$(\mathcal{G}, x, y, O)$}
        \STATE \textbf{continue} \COMMENT{Prune if no support from below}
    \ENDIF
    
    \IF{CanPlaceBlock$(\mathcal{G}, x, y, O)$}
        \STATE $\mathcal{G} \gets \text{ApplyPlacement}(\mathcal{G}, B, x, y, O)$
        \STATE $\mathcal{U} \gets \mathcal{U} \cup \{B\}$
        
        \IF{SolveRecursive$(\mathcal{I}, \mathcal{G}, \mathcal{U}) =$ \textbf{true}}
            \RETURN \textbf{true} \COMMENT{Propagate success upstream}
        \ENDIF
        
        \STATE \COMMENT{Backtrack: Revert state if path or stability fails}
        \STATE $\mathcal{G} \gets \text{RemovePlacement}(\mathcal{G}, B)$
        \STATE $\mathcal{U} \gets \mathcal{U} \setminus \{B\}$
    \ENDIF
\ENDFOR
\ENDFOR
\RETURN \textbf{false} \COMMENT{Exhausted branches without success}
\end{algorithmic}
\end{algorithm}



\subsection{Search Criteria}
To evaluate a candidate grid configuration $\mathcal{G}$, the solver verifies two primary conditions:
\begin{itemize}
    \item \textbf{Static Stability:} This condition is satisfied if every placed inventory block conforms to the convex hull stability constraint (Section~\ref{sec:support}), verifying that no component of the structural assembly will tilt or collapse under gravity. Integrating this quantitative physics metric ensures that the qualitative model accurately respects the domain's physical constraints.
    \item \textbf{Path Connectivity:} This condition is satisfied if the pathfinding module successfully extracts a continuous route from the Knight's starting location $K$ to the Princess's destination cell $P$ based on sequential spatial affordances (Section~\ref{sec:cell_walk}).
\end{itemize}

\subsection{Search Protocol and Backtracking}
The solver employs a recursive backtracking approach to systematically explore the valid state space:
\begin{enumerate}
    \item \textbf{Sequential Placement:} Blocks are selected one-by-one from the inventory $\mathcal{I}$ and assigned a local coordinate position via the mapping $L: B \to (x, y)$ alongside an orientation choice $\mathcal{O}$. Each block can serve as the starting point and all block orientations are tested recursively. Each potential placement enforces spatial Exclusion Rules and Foundational Restrictions (Section~\ref{sec:BlockPlace}).
    
    \item \textbf{Inventory Exhaustion:} The recursive engine continues branching until every asset within the inventory set has been successfully allocated to the grid.
    
    \item \textbf{Global Validation:} Once all the inventory is used, the agent verifies the center-of-mass (COM) stability of the complete assembly. It is necessary to check stability only after placing all blocks, as sometimes counterweights can stabilize an initially unstable arrangement.
    
    \item \textbf{Path Verification:} If the final configuration is completely stable, the agent invokes the grid-walking graph to evaluate connectivity between $K$ and $P$ via sequential cell affordances (Section~\ref{sec:cell_walk}).
    
    \item \textbf{State Reversion:} If the agent fails to satisfy any of the criteria previously described, it backtracks to a previous state, clearing the cell occupancy for the most recent node allocation ($c_{x,y} \to 0$), and tests for other alternative branches.
\end{enumerate}

\subsection{Termination Criteria}
The search terminates and yields a complete solution when both static stability and path connectivity are satisfied. The search can continue further to find multiple solutions, depending on the specific requirements of the agent solver.

\section{Discussion and Future Work}
\label{sec:discussion}

This work presents a hybrid qualitative solver for the puzzle game \textit{Camelot Jr.}, which requires understanding of structural stability and spatial path connectivity. Our solver agent leverages a symbolic paradigm to investigate the game state using common-sense logic, formalizing environmental interactions through qualitative structural affordances. Because the game mechanics involve complex physical equilibria, we integrate this symbolic layer with a numerical center-of-mass (COM) stability logic. By fusing these qualitative and numerical ingredients, the agent successfully solves arbitrary configuration challenges within the game's problem space.

We envisage several use cases and extensions for this kind of solving methodology. First, the framework can be used to train human spatial reasoning, specifically regarding structural stability and path visualization. Because the agent's decision-making process is inspired by human-centric thinking, the system can serve as a core engine for an intelligent tutoring system, providing meaningful, human-like hints tailored to a player's current solution attempt. Experiments with students in STEM fields, engaging with such interactive teaching tools, can be an effective educational aid.

Second, the solver's qualitative behavior opens up opportunities to investigate the specific structural and combinatorial parameters that increase task difficulty from a human perspective. By tracking metrics such as placement options, inventory count, and orientation and pruning efficiency across various levels, we can develop metrics that quantify how humans navigate spatial path-building games. 

Building upon this foundational setup, we plan to extend this hybrid paradigm to other physical puzzle games of a similar nature. Investigating how qualitative action affordances and rigorous physical constraints interact across different game environments will help uncover more generalized logics of human-like game-playing and tutoring, providing a clear roadmap for future inquiry.
\section*{Acknowledgments}
We acknowledge the funding by the Wallenberg AI, Autonomous Systems and Software Program (WASP) awarded by the Knut and Alice Wallenberg Foundation, and the funding by
\textit{Stiftelserna J.C. Kempes och Seth M. Kempes minne}, Sweden.

\bibliographystyle{named}
\bibliography{Spatial-references}

\end{document}